\documentclass{article}

\usepackage[preprint]{neurips_2026}

\usepackage[utf8]{inputenc}
\usepackage[T1]{fontenc}
\usepackage{hyperref}
\usepackage{url}
\usepackage{booktabs}
\usepackage{amsmath,amssymb,amsfonts}
\usepackage{nicefrac}
\usepackage{microtype}
\usepackage{xcolor}
\usepackage{graphicx}

\title{Small edits, large models:\\How Wikipedia advocacy shapes LLM values}

\author{%
  Jasmine Brazilek\thanks{Equal contribution.}\\
  Compassion Aligned Machine Learning (CaML)\\
  \And
  Maria Navas\footnotemark[1]\\
  Independent researcher\\
  \And
  Alexa Gnauck\\
  Pro-Animal Wikipedians (PAW)\\
}

\begin{document}
\maketitle

\begin{abstract}
Can a small group of volunteers shape how AI systems discuss animal welfare, just by editing Wikipedia? We show that they can. Wikipedia appears in nearly every major language model training corpus and is weighted more heavily than web-crawled text. The Pro-Animal Wikipedians (PAW), advocates who add sourced animal welfare content to relevant articles, have made 125 edits across 115 pages. We trace the influence of these edits on language models with gradient-based data attribution (Bergson, \citealt{lucia2026bergson}; MAGIC, \citealt{ilyas2025magic}). With TrackStar retrieval attribution on Llama 3.1 8B, PAW-edited sections make up 68\% of the highest-attributed documents for animal welfare queries ($p < 0.0001$), but only 52\% for unrelated queries about the same entities ($p = 0.53$). The model links PAW content specifically to animal welfare. MAGIC counterfactual influence estimation on Llama 3.2 1B, repeated across five random training-order seeds, shows the same pattern more sharply: in every seed, the ten most influential documents for animal welfare queries are all PAW edits, while the same ranking for general queries sits at chance. Mean PAW influence exceeds mean control influence at $p < 0.0001$ in every seed, an effect 6--30$\times$ larger than on general queries, and leave-subset-out validation gives Spearman $\rho = 1.00$ across all ten runs. A fine-tuning ablation confirms the attributions are causal: training on PAW content cuts perplexity on held-out animal welfare text by 32\%, while control training only helps on control text. A small, coordinated Wikipedia editing campaign therefore measurably shapes how language models handle the topics those edits address.
\end{abstract}

\section{Introduction}

Language models are becoming a primary way people get information. For advocacy organizations, this raises a practical question: can you influence what these models say about your cause? We show that Wikipedia editing is one way to do it. Wikipedia appears in nearly every major training dataset used to build language models (The Pile, RedPajama, Dolma, and others) and is given more weight than web-crawled sources because of its quality and breadth \citep{gao2020pile,soldaini2024dolma,weber2024redpajama}. This means that what Wikipedia says about a topic feeds directly into what language models say about it.

This creates an opportunity. Wikipedia has always been a place where small groups of dedicated editors shape how topics are presented \citep{yasseri2012dynamics}. Organized editing campaigns by civil society groups, political movements, and PR firms are common, ranging from state-backed takeovers of entire language editions to corporate reputation management \citep{distaso2013,kharazian2024governance}. What has gone mostly unnoticed is that these editing efforts now have a second effect: they shape the training data of language models, which in turn shape how millions of people get information. For advocacy organizations with limited budgets, this raises a concrete question: does editing Wikipedia actually change what AI systems say?

We study this phenomenon in the domain of animal welfare. The Pro-Animal Wikipedians (PAW) have conducted sustained editing of Wikipedia articles related to the use and perceptions of animal exploitation in articles about fast-food, animal sentience and politics. Using revision histories, we isolate the textual changes attributable to this group and quantify their downstream influence on language model behavior using gradient-based data attribution.

We employ Bergson \citep{lucia2026bergson}, an open-source library implementing TrackStar \citep{chang2024scalable}, to trace how individual Wikipedia edits influence model predictions. TrackStar works at the scale of billion-parameter models and corpora over 160 billion tokens, letting us go beyond surface-level keyword overlap and measure whether specific edits actually shift what a model says about animal welfare. This matters because the training examples that most influence a model's outputs are often not the ones that most obviously contain the relevant information \citep{chang2024scalable}.

\subsection{Why this matters for advocacy}

Animal advocacy organizations have limited resources and must choose their interventions carefully. Lobbying, public campaigns, and corporate outreach are familiar tools, but the rise of language models has created a new channel of influence that most organizations have not yet explored. If AI systems shape public understanding of animal welfare, then the data those systems learn from becomes a lever for change, and Wikipedia, as the most trusted and heavily weighted text source in model training, is the most accessible lever available. PAW is a loose coalition of editors who add sourced information about animal exploitation to relevant Wikipedia articles. Their edits are factual and follow Wikipedia's editorial policies, but they deliberately foreground animal welfare in articles about fast-food chains, animal sentience, and related topics.

This matters because LLMs trained on Wikipedia do not simply memorize articles; they learn associations, framings, and the relative salience of topics from the distribution of text they see. If a Wikipedia article on a fast-food chain includes a section on animal welfare controversies, models trained on that article learn to associate the brand with those controversies. If that section is absent, the association is weaker or missing. The question is whether these editorial choices have a measurable downstream effect, and if so, what kind of content drives it.

The stakes are real. Billions of people will use AI-powered systems in the coming years, and those systems will shape how they think about food, farming, and animals. If 125 Wikipedia edits by a handful of volunteers can measurably shift how a language model handles animal welfare, that is one of the cheapest and most scalable interventions available to the advocacy community: no technical expertise required, no API access, no budget beyond the time of committed editors. This paper measures that signal for the first time and confirms it exists.

\subsection{Contributions}

We make three contributions. First, we show for the first time that a real Wikipedia editing campaign (125 edits by a small group of volunteers) measurably and selectively influences language model behavior on the topics those edits address. Second, we confirm this with three independent methods (retrieval attribution, counterfactual training influence, and fine-tuning ablation), each showing that the effect is specific to animal welfare content and does not spill over to unrelated topics. Third, we frame Wikipedia editing as a practical, low-cost intervention for advocacy organizations: one that requires no technical infrastructure, no model access, and produces effects that carry through automatically whenever models are retrained on Wikipedia.

\section{Related Work}

This work sits at the intersection of three areas: data attribution for language models, Wikipedia as a site of organized editing, and the growing body of work on how training data affects model behavior.

Data attribution methods try to answer the question: which training examples caused a model to produce a given output? Influence functions \citep{koh2017understanding} were the first practical approach but are too expensive to run on large language models. TrackStar \citep{chang2024scalable} solved this by computing gradient similarity between training documents and queries, making attribution feasible for billion-parameter models over large corpora. MAGIC \citep{ilyas2025magic} goes further by backpropagating through the full training process to estimate counterfactual influence: what would change if a document were removed from training? Both are implemented in the Bergson library \citep{lucia2026bergson}. Other approaches include TRAK \citep{park2023trak} and the data-subset method of \citet{ilyas2022datamodels}, but these require training hundreds or thousands of models, making them impractical at scale. We chose TrackStar and MAGIC because they give complementary evidence (one measures similarity, the other measures causation) within a single tool.

Wikipedia is a standard component of language model training data. It appears in The Pile \citep{gao2020pile}, RedPajama \citep{weber2024redpajama}, Dolma \citep{soldaini2024dolma}, and the training data for GPT-3 \citep{brown2020language} and LLaMA \citep{touvron2023llama}, among others, and is given extra weight in most of these. Meanwhile, research on Wikipedia itself has long documented how its content is shaped by organized groups. \citet{yasseri2012dynamics} mapped edit wars on controversial topics. \citet{kharazian2024governance} showed how a small group captured governance of the Serbo-Croatian editions. Corporate PR editing is widespread \citep{distaso2013}. These two facts (Wikipedia is a key training source, and its content is shaped by organized editing) have not been connected empirically before this work.

A related line of work studies how training data composition affects model behavior. Data poisoning research \citep{wallace2021concealed,carlini2023poisoning} shows that injecting small amounts of manipulated data can steer model outputs. Our setting is different: PAW's edits are not adversarial injections. They are factual, policy-compliant Wikipedia contributions that foreground animal welfare. We show that this kind of legitimate, values-driven editing, not just adversarial attacks, shapes what models learn, which matters for how advocacy organizations think about influencing AI.

\section{Methods}

\subsection{Dataset Construction}

PAW's tracked editing activity spans 125 edits across 115 Wikipedia articles. For the attribution experiments we needed within-article pairing: each AW section matched to a non-AW section from the same page, controlling for article popularity, writing quality, and topic domain. Of the 115 pages, 31 contained both a clearly delineated AW section and a suitable non-AW control section of comparable length (e.g., company history, financial performance, store locations). Some articles lacked a non-AW section entirely (the PAW edit constituted most of the article's content), and others had no cleanly separable AW passage (the edit was an insertion within a mixed-topic section). From the 31 qualifying pages we drew 36 pairs (some pages yielded multiple distinct AW sections, such as Supermarket E with separate entries on cage-free eggs, sustainability, and animal welfare). This produced a balanced dataset of 72 sections.

For MAGIC and the fine-tuning ablation, within-article pairing is not required because these methods measure a different quantity: how much a document's presence in training causally affects query loss. The control set need only represent typical Wikipedia text without animal welfare content. We therefore extracted AW text from all 125 tracked edits across 115 pages; after removing 7 entries with empty content fields, this yielded 118 AW sections from 108 unique pages. We paired these with 118 random chunks from WikiText-103 (seed 42) as controls, producing a 236-document dataset. Using unrelated Wikipedia text rather than pre-PAW versions of the same articles is deliberate: pre-PAW content from the same pages would share entities and partial topic overlap with the AW sections, weakening the separation between conditions. Our design guards against confounds from text quality or length differences: if controls were simply worse text, AW documents would dominate attribution for all queries, not just animal welfare queries.

Two query sets were used to probe attribution: 80 animal welfare queries (e.g., ``What is Supermarket D's animal welfare policy?'') and 90 general queries about the same entities (e.g., ``How many stores does Supermarket D have?''). The general queries serve as a negative control: if attribution signals are genuinely topic-specific, AW content should not be preferentially attributed to queries that mention the same companies but ask about unrelated matters.

\subsection{TrackStar Retrieval Attribution}

TrackStar \citep{chang2024scalable} computes gradient similarity between training documents and queries to estimate retrieval-based attribution. For each query, it ranks every training document by how strongly the document's training gradient aligns with the query's gradient, producing a score that reflects semantic relevance as learned by the model. We ran TrackStar via Bergson on Llama 3.1 8B using the paired 72-document dataset against both query sets.

\subsection{MAGIC Training Influence}

MAGIC goes beyond retrieval similarity by backpropagating through the full training history. It fine-tunes a model on the document set, records optimizer checkpoints at each step, then traces backwards to compute how much each document's gradient contributed to the model's loss on a given query. The resulting score estimates the counterfactual: how would query loss change if this document were removed from training? Bergson's MAGIC implementation reports per-document scores in original-doc order with the convention that more-negative scores indicate documents whose removal would increase query loss most (i.e., the most influential documents).

We fine-tuned Llama-3.2-1B with LoRA (rank 32, targeting q\_proj and v\_proj, alpha 64) on the 236-document dataset (118 PAW edits + 118 WikiText-103 controls) for one epoch with a polynomial learning rate schedule (peak lr $= 4 \times 10^{-4}$, 25\% warmup), AdamW optimizer ($\beta_1 = 0.95$, $\beta_2 = 0.975$), weight decay 0.01, batch size 4, and fp32 precision.

\textbf{Multi-seed protocol.} The training-order shuffle is seeded; per-document MAGIC scores depend on the order documents appear during fine-tuning, and pilot work on a separate experiment indicated that single-seed estimates can carry seed-dependent noise that internal validation does not flag. We therefore ran the full pipeline at five random seeds (1, 7, 42, 99, 256) for each of the two query sets (10 runs total). Each run included its own leave-subset-out validation with 5 subsets; all 10 runs returned Spearman $\rho = 1.00$ ($p < 10^{-23}$), with score variances in the same order of magnitude across seeds and no numerical instability.

We report aggregate findings (top-$k$ composition and mean-difference tests) pooled across all five clean seeds, and we treat any individual document's specific rank as supported only when it replicates in a substantial fraction of seeds.

\subsection{Fine-Tuning Ablation}

To test whether different training content produces measurable differences in model behavior, we fine-tuned two separate Llama-3.2-1B models using the same LoRA configuration: one on the 118 AW sections and one on the 118 control sections. We then measured how well each model predicted both text sets, testing whether each model performs better specifically on the type of content it was trained on.

\section{Results and Discussion}

All three methods point to the same conclusion: PAW's Wikipedia edits influence how language models handle animal welfare topics, with no effect on unrelated queries about the same companies.

\subsection{TrackStar Retrieval Attribution}

We applied Bergson's TrackStar method to a balanced dataset of 72 Wikipedia sections (36 animal welfare, 36 control sections paired from the same articles) and measured retrieval-based attribution scores across 80 animal welfare queries and 90 general queries.

For animal welfare queries, AW sections were overrepresented among the highest-scoring documents: 68.0\% of top-5 results ($p < 0.0001$, one-sample $t$-test against the 50\% base rate), 66.6\% of top-10 ($p < 0.0001$), and 63.7\% of top-15 ($p < 0.0001$). In 64 of 80 queries, AW content dominated the top-5. Mean attribution scores were higher for AW sections (0.549) than control sections (0.444; Mann-Whitney $U$, $p < 0.0001$).

To ensure that the observed influence was specific to the domain of animal welfare and not a systemic bias of the attribution method, we conducted a control experiment using the 90 general-topic queries. For these control queries, the representation of PAW-edited documents in the top ranks did not significantly deviate from the 50\% baseline (Top-5: 51.7\%, $p=0.528$; Top-10: 49.4\%, $p=0.772$; Top-15: 48.3\%, $p=0.227$; one-sample t-test). This contrast confirms that the high influence scores recorded for PAW editions are content-dependent and specific to animal welfare topics, and rules out the possibility that the signal comes from PAW sections simply being longer or better written.

\begin{figure}[t]
    \centering
    \includegraphics[width=0.8\textwidth]{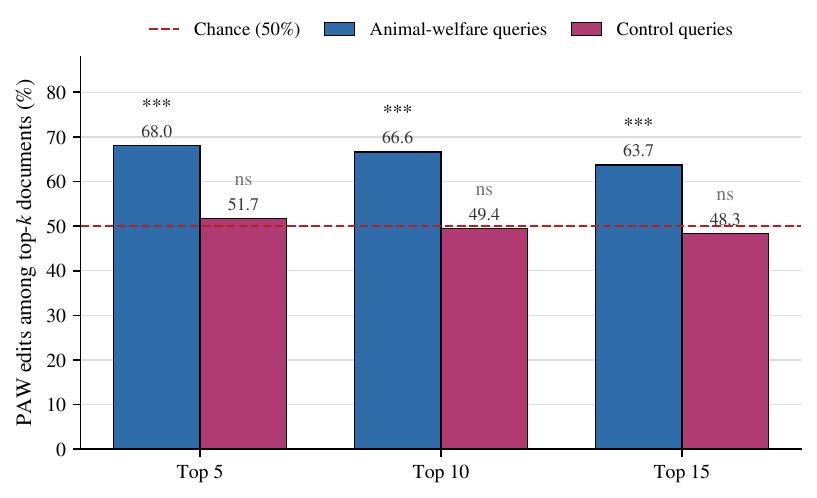}
    \caption{Domain-specific influence of PAW Wikipedia edits. Share of PAW-edited sections among the top-$k$ TrackStar-attributed training documents for Llama 3.1 8B. PAW edits are significantly overrepresented for animal welfare queries (binomial test against the 50\% baseline; ***: $p < 0.0001$) and indistinguishable from chance for control queries about the same entities (ns: $p > 0.05$). The dashed line marks chance level (50\%).}
    \label{fig:trackstar}
\end{figure}

\subsection{MAGIC Training Influence Attribution}

MAGIC traces influence through the full training process, estimating how much each document causally contributes to the model's predictions. We fine-tuned Llama-3.2-1B on the 236-document dataset (118 PAW edits + 118 WikiText-103 controls) at five random seeds for each of the two query sets, then pooled the results.

\begin{figure}[t]
    \centering
    \includegraphics[width=\textwidth]{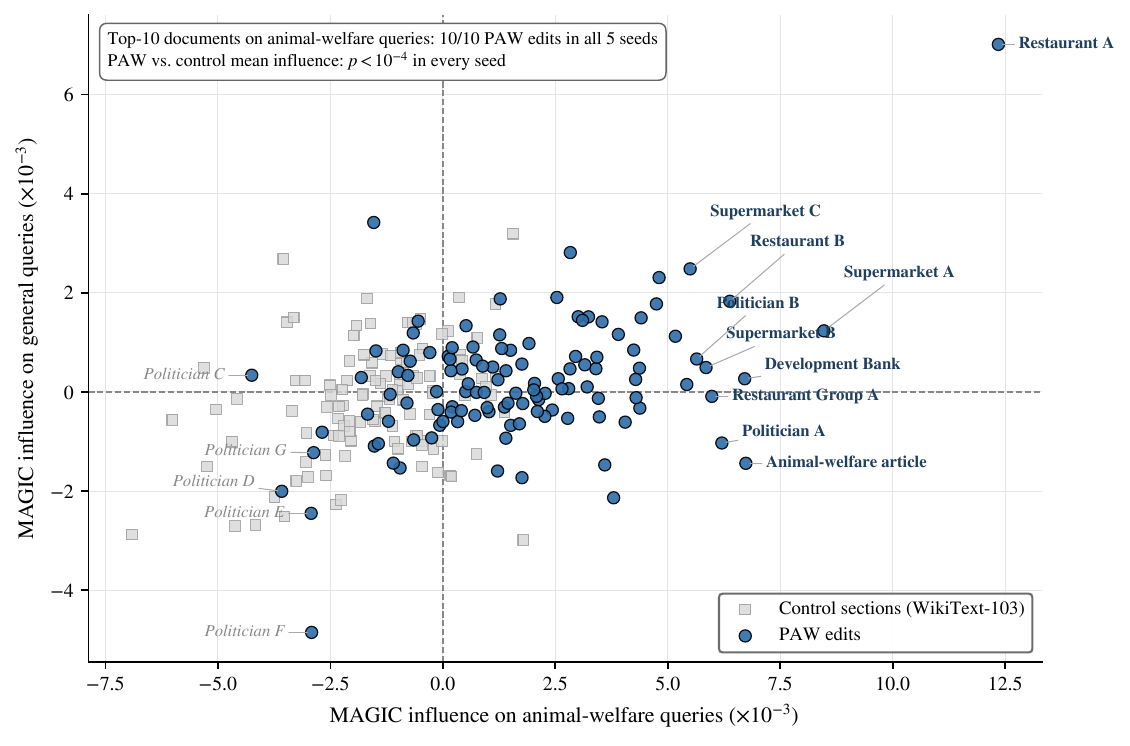}
    \caption{MAGIC counterfactual influence per training document, averaged across five random training-order seeds. Each point is one of the 236 training documents; the $x$-axis is its mean influence on animal welfare queries and the $y$-axis is its mean influence on general queries about the same entities (sign-flipped so that larger means more influential under the Bergson convention; axis units are $10^{-3}$). Blue circles are PAW edits, light squares are WikiText-103 controls. Most PAW edits sit to the right of zero on the $x$-axis but cluster near zero on the $y$-axis: their influence is topic-specific to animal welfare. Labelled top-influence PAW edits (in bold) are concrete, dated welfare commitments and legislative records (Restaurant A, Supermarket A, Restaurant B, Restaurant Group A, Supermarket B, Supermarket C, the Development Bank, a national animal-welfare article, and the records of Politician A and Politician B). The least-influential PAW edits (italics, lower left) are mostly pages of individual politicians whose animal-welfare sections contain brief position statements rather than concrete commitments. Across all five seeds, the top-10 most influential documents on AW queries are 10/10 PAW edits in every seed; on general queries the same top-10 sits at the 50\% chance baseline.}
    \label{fig:magic_scatter}
\end{figure}

\textbf{Top-10 dominance is total and replicates across every seed.} For each seed we ranked all 236 training documents by MAGIC influence on the 80 animal welfare queries (most-negative score = most influential in the bergson convention). In \emph{every one of the five seeds}, all 10 of the top-10 most influential documents were PAW edits (Figure~\ref{fig:magic_scatter}). On the 90 general queries about the same entities, that figure dropped to 4--6 of the top-10, right at the chance baseline of 5/10, in every seed. The 100\% versus chance split for AW vs.\ general queries replicates without exception across all five training-order shuffles.

\textbf{Aggregate score difference is significant in every seed.} Mean MAGIC influence was significantly more negative (more influential) for PAW edits than for WikiText-103 controls on AW queries in all five seeds (Mann-Whitney $U$, $p < 0.0001$ each seed; mean difference $-0.0028$ to $-0.0039$). On general queries the same comparison gave mean differences of $-0.0001$ to $-0.0006$, an effect 6--30$\times$ smaller, and reached $p < 0.05$ in only 4 of 5 seeds. The asymmetry the TrackStar analysis showed at the retrieval level reappears at the counterfactual-training-influence level: PAW edits influence the model specifically on animal welfare topics, not on the entities in general.

\textbf{Internal validation.} Leave-subset-out validation (5 subsets per run) returned Spearman $\rho = 1.00$ ($p < 10^{-23}$) for all 10 runs, with no numerical instability and consistent score-magnitude ranges across seeds.

\textbf{Which PAW documents are most influential?} Specific document rankings within the top-10 vary modestly across seeds (this is the part of the MAGIC signal most sensitive to training-order shuffles), but several PAW edits appear consistently. Restaurant A's section describing its parent group's battery-cage-egg phase-out commitment and an animal-welfare NGO's subsequent criticism for slow progress was in the top-10 on 4 of 5 seeds. Politician H's record on a state farm-animal-confinement ballot measure and a federal farm bill was in the top-10 on 3 of 5 seeds, as was a national animal-welfare committee's pig-farm review section, the Development Bank's section on activist criticism for financing intensive animal agriculture, and Supermarket A's 2027 cage-free egg commitment. The common thread across these top-replicating PAW edits is concrete, dated detail: specific commitments, specific organizations doing the criticizing, specific policies, specific years. Edits without that texture, such as scorecard ratings or brief political position statements, did not consistently rank near the top.

\subsection{Fine-Tuning Ablation}

The ablation produced a clean separation. The model trained on PAW content had a perplexity of 8.4 on animal welfare text, compared to 12.4 for the model trained on control content: it was 32\% better at predicting PAW-style text. Going the other direction, the control-trained model had a perplexity of 11.4 on control text versus 16.1 for the PAW-trained model: 29\% better on its own domain. Against the unmodified base model (perplexity 11.9 on AW text, 15.5 on control text), each fine-tuned model improved on its own training content while leaving the other essentially unchanged. This confirms that the attribution signal MAGIC detects is real: training on animal welfare content specifically improves how the model handles animal welfare, and training on control content specifically improves how it handles control text.

\begin{figure}[t]
    \centering
    \includegraphics[width=0.9\textwidth]{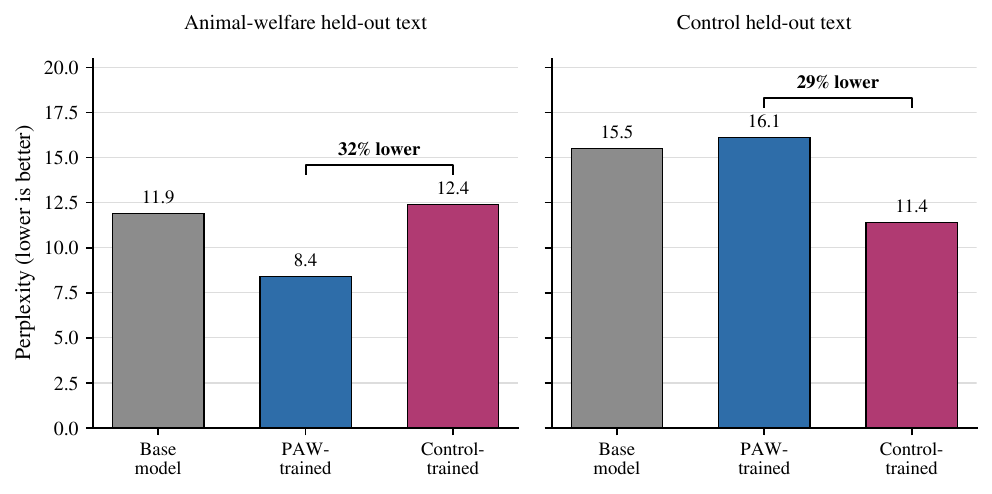}
    \caption{Fine-tuning ablation. Each model performs better on the type of text it was trained on. The PAW-trained model is 32\% better (lower perplexity) on animal welfare text; the control-trained model is 29\% better on control text.}
    \label{fig:ablation}
\end{figure}

\section{Limitations}

\textbf{Model scale.} Our attribution experiments used Llama 3.1 8B (TrackStar) and Llama 3.2 1B (MAGIC, ablation). These are smaller than models like GPT-4 or Claude, but the key point is that those frontier models were also trained on Wikipedia, meaning PAW's edits are already in their training data. We demonstrate the mechanism at scales where attribution is computationally feasible; the next section argues that the effect persists, and likely strengthens, at frontier scale.

\textbf{Dataset coverage.} The MAGIC and ablation experiments used all 125 PAW edits (118 after cleaning); TrackStar used a 36-pair within-article subset. Future work with topic-matched controls, that is, animal welfare text not written by PAW, could tease apart whether the signal comes from PAW's specific framing or from animal welfare content more broadly.

\textbf{What MAGIC measures.} We measure influence on model prediction quality, not on what a chatbot actually says in conversation. Bridging that gap is important future work.

\textbf{Seed sensitivity of specific document rankings.} The MAGIC results in this paper are pooled across five random training-order seeds with internal validation $\rho = 1.00$ on every run, so the headline aggregate findings (AW vs.\ control top-10 share, mean influence difference, asymmetry between AW and general queries) are not seed-dependent. Specific rankings of individual PAW documents within the top-10 do vary modestly across seeds, and we report only the documents that replicate in at least 3 of 5 seeds.

\section{Why This Already Applies to Frontier Models}

GPT-4, Claude, Gemini, and every other frontier model trained on web data has already seen Wikipedia, including the pages PAW edited. The influence we measure here is not hypothetical; it is already baked into the models people use today. The only question is whether the signal gets diluted below a meaningful threshold in a much larger training corpus. We argue it does not.

In theory, a single document's influence shrinks with total training data, roughly $O(1/N)$ where $N$ is the number of tokens. But several factors work against simple dilution:

First, Wikipedia is upweighted in every known pretraining mixture. The Pile applies a ${\sim}3\times$ weight; RedPajama and Dolma use $2$--$5\times$; and multi-epoch training means Wikipedia text may be seen 3--4 times across training. If we denote the upweighting factor as $w$ and the number of epochs as $e$, the effective contribution scales as $O(w \cdot e / N)$. For a frontier model trained on 15T tokens with $w = 3$ and $e = 3$, the dilution relative to our 1B-model setup (trained on ${\sim}1$T tokens with $w \cdot e \approx 1$) is roughly $15\text{T} / (3 \cdot 3 \cdot 1\text{T}) \approx 1.7\times$, far less than the $15\times$ that the raw token ratio suggests.

Second, the information PAW edits introduce is not unique to Wikipedia. The same corporate welfare commitments, backtracked pledges, and NGO criticisms appear in news coverage, industry trade publications, corporate press releases, and advocacy reports, all of which end up in Common Crawl and thus in pretraining corpora. This means PAW edits are reinforced by corroborating text from other sources. A model trained on 15 trillion tokens has seen a major grocery chain's cage-free backtrack not only on Wikipedia but in industry trade publications and animal-welfare NGO press releases. Rather than a single signal that dilutes with scale, the PAW-edited content sits within a broader cluster of semantically similar documents that collectively push the model toward the same associations. Larger models have more capacity to retain these per-document associations: empirical work shows that larger models memorize more training examples, not fewer \citep{carlini2023poisoning}. The practical effect is that Wikipedia's role shifts from sole source to authoritative anchor within a reinforcing cluster.

Third, PAW edits are not isolated documents; they modify articles that already contain substantial context about the same entities. The model encounters the PAW-edited content embedded within pages about Restaurant C, Restaurant E, Supermarket E, etc., creating associations between these high-traffic entities and animal welfare framings. This contextual embedding means the edits influence not just direct recall but the broader associative structure of the model's knowledge.

Taken together, the real dilution from our experimental setup to frontier models is on the order of $2$--$5\times$, not the $15$--$30\times$ that raw data volume would suggest. Our measured effects are large enough that a $2$--$5\times$ reduction still leaves a meaningful signal. And since frontier models have already been trained on the Wikipedia pages PAW edited, that signal is already present in the models millions of people use daily. The practical implication is immediate: PAW's 125 edits have already shaped how frontier language models handle animal welfare topics. Scaling up to several hundred well-placed edits (well within the capacity of a small volunteer group) would strengthen this influence further with each new model training run.

\section{Acknowledgements and Disclosures}

This research was conducted at Compassion Aligned Machine Learning (CaML). We thank the Pro-Animal Wikipedians for providing access to their editing records and for their sustained work improving animal welfare coverage on Wikipedia. Compute for the MAGIC and ablation experiments was provided via RunPod. The Bergson library by Quintin Lucia and Nora Belrose at EleutherAI made the data attribution experiments possible.

Disclosure: The authors are affiliated with animal welfare research organizations. PAW's Wikipedia edits are factual, policy-compliant contributions that have survived standard editorial review. This paper documents the downstream influence of those edits on language models; it does not advocate for any particular editorial content. The fine-tuned models and the underlying edit dataset are available from the authors on request.

\bibliographystyle{apalike}
\bibliography{references}

\appendix
\section{Query Examples and Data Availability}

The fine-tuned models, the underlying datasets, and the query sets are available from the authors on request. Below we list example queries from each set.

\subsection{Animal Welfare Queries (80 total)}

These queries ask about animal welfare policies, practices, or controversies related to specific entities covered by PAW edits. Examples:

\begin{itemize}
    \item ``What is Restaurant C's chicken welfare policy?''
    \item ``Supermarket D animal welfare''
    \item ``Restaurant F animal welfare''
    \item ``Company A animal testing''
    \item ``Supermarket E cage free eggs''
    \item ``Restaurant E animal welfare controversies''
    \item ``What are the animal welfare concerns with Restaurant G's supply chain?''
    \item ``Restaurant H animal welfare''
    \item ``Politician H Animal Welfare''
    \item ``Restaurant I animal welfare policy and practices''
\end{itemize}

\subsection{General Queries (90 total)}

These queries mention the same entities as the AW queries but ask about unrelated topics (company size, history, products). They serve as a negative control: if attribution is truly topic-specific, these queries should not preferentially retrieve animal welfare content. Examples:

\begin{itemize}
    \item ``How many stores does Supermarket D have?''
    \item ``Restaurant F''
    \item ``Restaurant C franchise model''
    \item ``Supermarket E store locations''
    \item ``Restaurant E history and founding''
    \item ``Restaurant H revenue 2024''
    \item ``Restaurant G sandwich menu''
    \item ``Restaurant I number of employees''
\end{itemize}

The fine-tuned model weights are available on HuggingFace; the full query sets and both document datasets (72-section paired set and 236-document full set) are available from the authors on request.

\end{document}